\documentclass[twocolumn]{article}

\usepackage[utf8]{inputenc}
\usepackage{titling}
\usepackage{authblk}
\usepackage{graphicx}
\usepackage{microtype}
\usepackage{csquotes}
\usepackage[backend=biber, style=numeric]{biblatex}
\usepackage[acronym, nomain]{glossaries}
\usepackage{xspace}

\makeglossaries
\newacronym{ppr}{PPR}{Product-Process-Resource}
\newacronym{fpd}{FPD}{Formalized Process Descriptions}
\newacronym{css}{CSS}{Capability-Skill-Service}
\newacronym{mes}{MES}{Manufacturing Execution System}
\newacronym{rdf}{RDF}{Resource Description Framework}
\newacronym{owl}{OWL}{Web Ontology Language}
\newacronym{opc}{OPC UA}{Open Platform Communications Unified Architecture}
\newacronym{ass}{ASS}{Asset Administration Shell}
\newacronym{soa}{SOA}{Service-Oriented Architecture}
\newacronym{mtp}{MTP}{Module Type Package}

  \author[1]{Aljosha Köcher}
  \author[2]{Alexander Belyaev}
  \author[3]{Jesko Hermann}
  \author[4]{Jürgen Bock}
  \author[5]{Kristof Meixner}
  \author[6]{Magnus Volkmann}
  \author[7]{Michael Winter}
  \author[8]{Patrick Zimmermann}
  \author[9]{Stephan Grimm}
  \author[2]{Christian Diedrich}
  
  \affil[1]{Helmut-Schmidt-Universität, aljosha.koecher@hsu-hh.de}
  \affil[2]{Otto von Guericke Universität Magdeburg}
  \affil[3]{Technologie-Initiative SmartFactory KL e.V.}
  \affil[4]{Technische Hochschule Ingolstadt}
  \affil[5]{Christian-Doppler-Laboratory-SQI, TU Wien}
  \affil[6]{TU Kaiserslautern, WSKL}
  \affil[7]{RWTH Aachen University}
  \affil[8]{Fraunhofer-Institut für Gießerei-, Composite- und Verarbeitungstechnik IGCV}
  \affil[9]{Siemens AG}
  
  \title{A Reference Model for Common Understanding of Capabilities and Skills in Manufacturing 
  \\
  \vspace{5mm}
  \large
  Terms and Definitions for a Standardized Model of Capabilities, Skills and Services
  \vspace{8mm}
  }

  \date{}
  
\begin{document}

  \begin{titlingpage}
	\maketitle
  	\abstract{
	In manufacturing, many use cases of Industry~4.0 require vendor-neutral and machine-readable information models to describe, implement and execute resource functions.
	Such models have been researched under the terms \emph{capabilities} and \emph{skills}.
	Standardization of such models is required, but currently not available.
	This paper presents a reference model developed jointly by members of various organizations in a working group of the \emph{Plattform Industrie 4.0}. This model covers definitions of most important aspects of capabilities and skills. It can be seen as a basis for further standardization efforts.
  
  	\vspace{10 mm}
  }
  \end{titlingpage}
  


\def\cssmodel{CSS~Model\xspace}

\section{Introduction}
\label{sec:introduction}

A substantial trend in future manufacturing is the requirement for a faster reaction to market uncertainties resulting in a need for more flexibility in industrial production~\cite{Pfrommer2013}. 
This flexibility concerns many different aspects, e.g., the ability for fast introduction of new products or product variants. Therefore, the possibility to efficiently produce high-mix scenarios under low-volume down to lot-size-1 is necessary, which requires new concepts for production control as well as the ability to react to problems and disturbances within production and supply chains. 
One possible approach to tackle this challenge is modularizing production resources and requirements and abstracting to the functions provided or requested.
However, the information on available and provided functions must be communicated to the interaction partners so that all interaction partners understand them similarly. 

In recent years, many research activities have taken place to develop the required concepts in detail and elaborate their application in the industrial domain.\footnote{For instance, the Conference on Emerging Technologies and Factory Automation hosts a special session on this topic.}
Accordingly, a wide variety of terms and concepts have emerged that are relevant for describing functionalities of assets~\cite{Froschauer2022}.

In today's discussion in the I4.0 community, various similar terms, e.g., ``capability'', ``skill'' or ``task'', are used to describe a function of an asset.
However, to the best of our knowledge, a clean differentiation of these terms does not exist. 
Often, different authors use various terms to describe similar concepts.Vice versa, the same term is assigned to contrary concepts. Furthermore, functions need to be communicated on different levels of abstraction for use cases ranging from automated order handling in supply chains to manufacturing execution. 

All these issues complicate the comparison of existing approaches and lead to confusion. Accordingly, the emerging solutions cannot be interoperable per se. 
This work aims to consolidate the different company- and institution-specific terms by answering the following research questions.
\begin{itemize}
    \item What is the essential set of concepts from flexible production that must be covered to communicate available and required functions?
    \item What does a model look like that relates these concepts?
    \item What are potential technical implementations of this model?
\end{itemize}

As a result, the authors present the \gls{css} model which was jointly developed in a working group of \emph{Plattform Industrie 4.0}. The \gls{css} model can be seen as a reference model of capabilities, skills, and services, defining and categorizing them for a clear distinction and specifying their relationships with each other.
Furthermore, technology mappings for each aspect of the model are presented and a distinction of the \gls{css} model from existing, similar approaches is made.





The remainder of the paper is structured as follows:
Chapter~2 gives an overview of related activities and research work. 
Chapter~3 introduces the \gls{css} reference model with extended definitions and explanations of the concepts \emph{capability}, \emph{skill}, and \emph{service} before Chapter~4 presents potential technology-specific implementations. 
Chapter~5 summarizes approaches that pursue similar goals, i.e., the encapsulation of automation functions and description of their interfaces. 
Finally, Chapter~6 recaps and critically evaluates the \gls{css} reference model and provides an outlook of possible future research activities.

\section{Related Work}
\label{sec:relatedWork}

In production systems engineering, organizations work in multidisciplinary engineering environments, where stakeholders maintain different views on the manufactured products and the production system.
The \gls{ppr} concept, described in~\cite{schleipen2015requirements}, represents the three major aspects of production systems engineering.
\emph{Products} describe input and output products, \emph{processes} describe production processes required to transform input into intermediate and output products, and \emph{resources} describe production resources that execute the production processes.
The \gls{fpd}, defined in VDI~3682~\cite{vdi_3682}, provides a visual and formal model to describe these aspects.

Pfrommer~et. al~\cite{Pfrommer2013} introduced \emph{skills} as additional element to the \gls{ppr} concept.
The authors defined \emph{skills} as vendor-independent representations of production process functionality required by a product and provided by a resource.
These characteristics enable the abstraction between production processes and resources.

A more thorough terminological discussion of the notions of functionality and function with their relation to capabilities (and even capacities) can be found in \cite{BST_CapabilitiesCapacitiesandFunctionalities_2021}.
This work also includes pointers to further references on formal terminological clarification of \emph{function} in the ontology engineering community.

In earlier publications, the concepts of capabilities and skills were often used interchangeably~\cite{Froschauer2022}.
However there is a more apparent distinction between capabilities and skills in recent literature.
While capabilities are often defined as an abstract description of a function provided by a machine, skills are typically seen as executable implementations of these functions that might be used to execute a process on a machine.
Even though this distinction is slowly starting to emerge, there is no holistic integrated model of \gls{ppr} on one side and capabilities and skills on the other side so far~\cite{Froschauer2022}. 

The \textit{Plattform Industrie 4.0}~\cite{PlattformI40.2020b} describes the individual generation of processes based on product descriptions and defined capabilities. Capabilities are defined as a vendor-neutral description of functions, while a skill is an implementation of a resource to realize a function \cite{PlattformI40.2020b}. Possible technologies, e.g., to model capabilities in ontologies or different realizations of skills are discussed \cite{PlattformI40.2020b}. 
To this discussion around capabilities and skills comes another level of abstraction when talking about supply chains spanning across company borders.
For such a network with service providers to share production resources, \cite{Liu.2019b} defines the term \emph{cloud manufacturing} as ``a model for enabling aggregation of distributed manufacturing resources [\dots] to a shared pool of configurable manufacturing services that can be rapidly provisioned and released with minimal management effort or service operator and provider interaction''.

While standards around capabilities, skills and services are currently non-existent, a selection of standards is regularly incorporated into models. In addition to VDI guideline 3682, most often, standards for the definition of process types such as DIN~8580 (manufacturing processes) or VDI~2860 (handling) are used (e.g., \cite{KHV+_AFormalCapabilityand_9820209112020, WBS+_AnOntologybasedMetamodelfor_9820209112020}). Furthermore, the state machine defined in \emph{PackML}/ISA~88 is often used to model the behavior and interactions of skills (e.g., \cite{Dorofeev.2018, KHV+_AFormalCapabilityand_9820209112020}

\section{Capability-Skill-Service Model} 
\label{sec:model}

\begin{figure*}[ht]
    \centering
    \includegraphics[width=0.9\textwidth]{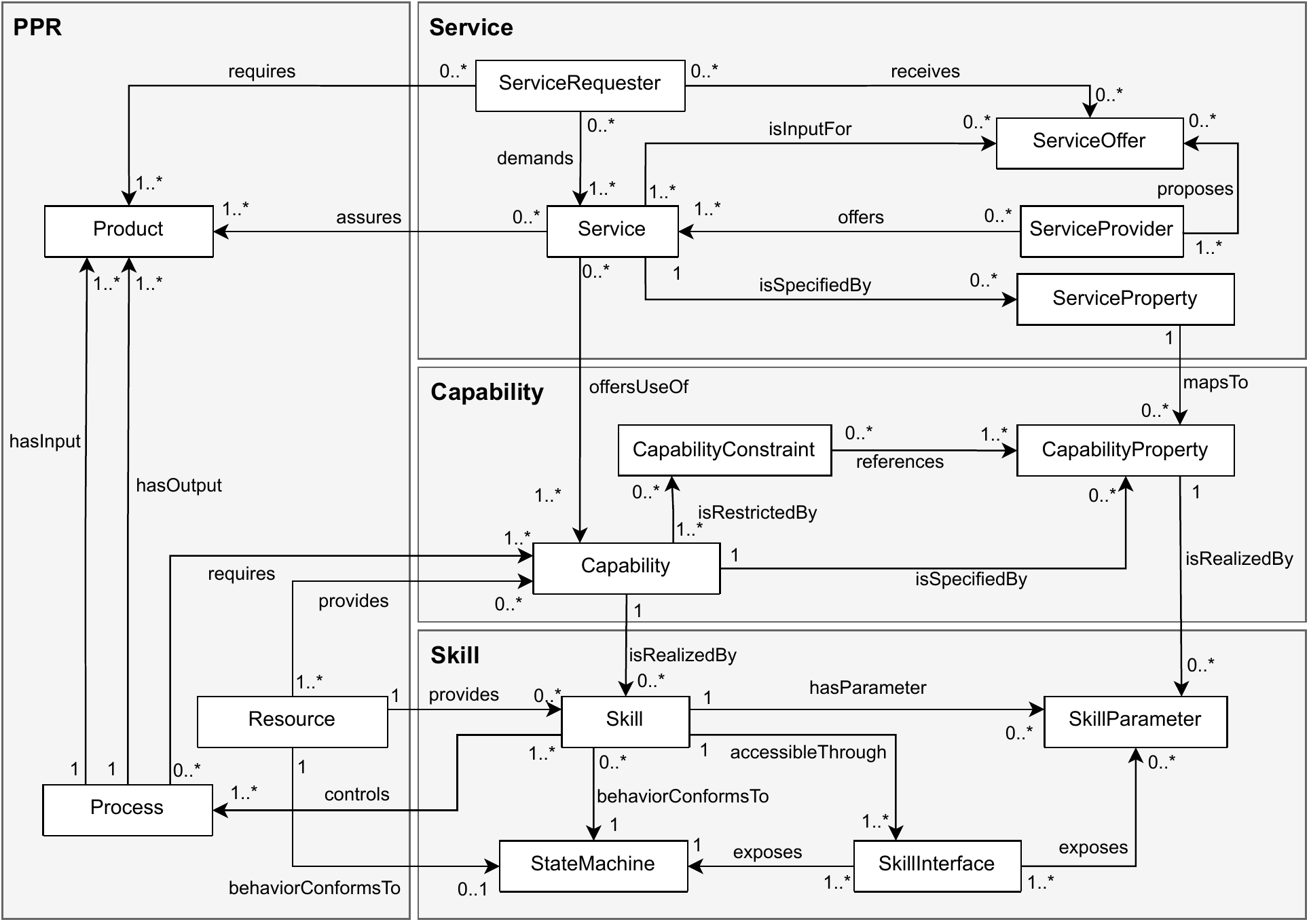}
    \caption{The reference model of capabilities, skills, and services with alignment to the PPR approach.}
    \label{fig:reference_model}
\end{figure*}

As an attempt to consolidate the terminology around capabilities and skills for manufacturing, the conceptual model presented in this section extends the well-established PPR paradigm by the additional notion of a \emph{manufacturing function}. Thereby, the otherwise rigid link between processes and resources is decoupled  by a separate description of required and provided functionality\footnote{We understand the notion of \emph{function} in a general sense, as e.g. discussed in the ontology engineering literature \cite{BST_CapabilitiesCapacitiesandFunctionalities_2021}, not in a mathematical or computer science sense -- the function of e.g. a hammer is to drive in nails.}.
However, as manufacturing systems engineering requires information modeling at different levels of system functionality, it is not sufficient to simply add \emph{function} as an element to PPR. 
Rather, the notion of function needs to be reflected at all the relevant levels, namely the level of invocation of automation components, the level of abstract description of an in-house factory's functionality, and the level of external offering of such functionality to form shared production networks.
To this end, we introduce three different model elements by which we extend PPR, all of which represent aspects of function for different uses. These are the \emph{skill} as invocable function, the \emph{capability} as abstract description of function, and the \emph{service} as a business-level offering of function to external partners.

After a short discussion of requirements and an overview of the model, the three terms \emph{capability}, \emph{skill} and \emph{service} are presented in separate subsections.

\subsection{Requirements \& Model Overview}
\label{sec:requirements}
An overview of requirements from various publications can be found in \cite{Froschauer2022}. Only the most relevant requirements for this work are discussed here since more concrete requirements, e.g., regarding solution technologies, are out of the scope of this conceptual model.
Models of capabilities, skills, and services need to foster more efficient approaches to production planning and production system reconfiguration. This can be considered a paramount requirement from which all others can be derived.
Both \emph{matchability} and \emph{executability} are often mentioned as a requirement highlighting the needed reflection of functions on different levels \cite{Froschauer2022}. While matchability may best be achieved with formal models, executability necessitates bindings to implementation technologies. Thus, a clear distinction between these concepts is needed. 
Skills must have a communication interface and individual skill states need to be expressed \cite{Froschauer2022}.

Figure~\ref{fig:reference_model} illustrates the developed \gls{css} reference model for which the previously discussed requirements were considered. 
The model distinguishes between the three aspects capability, skill, and service. In each of the aspects, the main concepts and their relations are defined. Furthermore, the model represents interfaces between the model elements. In addition, the main concepts of each aspect are related to \gls{ppr} concepts.

\subsection{Services}
\label{sec:services}
One of the promising and future-oriented developments of the manufacturing industry is the upcoming transformation of industrial production into shared production. According to this scenario, production sites will form cross-company networks.
In such networks, resources can offer their manufacturing capabilities and integrate the capabilities of external partners into their own production processes based on specific orders. 
Such scenarios imply automated order processing in supply chains spanning across various companies. The challenge in realizing such shared production networks is interoperability. 
This work addresses the question of whether it is sufficient to talk about the provided capabilities of resources in this context. The proposition put forward is that there can be important further parameters beyond pure technical description of provided functionalities.
These are, for example, information on economic criteria such as delivery dates, cost, and agreements regarding documentation or maintenance, certification, rating, etc. 
For modeling such issues and distinguishing them from the concept of capability, this work introduces the term \emph{service} in an economic sense which represents a set of technical capabilities supplemented by an organizational and economic description.
It should be noted that the service concept presented here is different from the same term used in information technology (see Section~\ref{sec:webservices} for a distinction). 

In the context of the \gls{css} model, a service requester, who provides a specification of a requested service with its properties, demands suitable services.
If a service provider can provide the demanded service, it may propose an offer as the basis of a binding contract to execute one or more services. The service requester can accept the proposal in a specified time period under the proposed conditions. If service requesters are searching for services through a marketplace, multiple service offers may be created by different service providers. These offers could be mutually exclusive, or could also be combined to fulfill a requested service.

\subsection{Capabilities}
\label{sec:capabilities}
We define a capability as ``an implementation-independent specification of a function in industrial production to achieve an effect in the physical or virtual world''. Thus, a capability describes a function in a production process. Usually, capabilities specify production functions that have an effect in the physical world. Nevertheless, software functions that only apply to the virtual world may also be modelled as a capability. Capabilities that specify a production function should refer to terms of an actual manufacturing method, such as ``drilling'', along with properties and constraints to describe constraints of the application. An example of a capability could read ``drilling a hole with a particular depth and a diameter into certain types of material''. 

Capabilities are either provided by production resources that claim the ability to apply the expressed function or are required by a process as part of a product's functional requirements. 
Required and provided capabilities typically differ, e.g., because provided capabilities are described in more detail to enable reuse for a wide variety of processes.
A matching between required and provided capabilities is thus necessary to find candidates for a suitable sequence of production steps for given requirements.
This matching can initially be done on a descriptive level – e.g., by comparing capability types and their properties – regardless of which actual resources execute these process steps later.

In the CSS model, capabilities are connected to the concepts \emph{skill} and \emph{service}. The implementation of capabilities is possible by means of skills, which contain details at the level of implementation and invocation of automation functions. In a broader supply chain network outside an internal production setup, capabilities are offered to stakeholders by means of services.

\subsection{Skills}
\label{sec:skills}
We define a skill as ``an executable implementation of an encapsulated (automation) function specified by a capability''. 
A skill is provided by a resource in the production environment and enables the realization of a capability. Every capability may reference multiple skills in the production environment that act as implementations for this capability. 
Skills must have a skill interface allowing external systems (e.g., a \gls{mes}) to interact with the provided function. 
Every skill's behavior needs to follow a standardized state machine describing possible states and transitions. The state machines needs to be exposed via the skill interface so that the current state can be monitored and transitions can be triggered. 
The execution of one or more skills allows to control production steps. 
Skills can have input or output parameters enabling the execution or monitoring of a skill. These parameters must also be exposed by the skill interface in order to set or get parameter values.
Specifying input parameters makes it possible to execute defined production steps with an individual configuration. 
Properties of a capability may refer to parameters of a related skill. Thus, the parameters can be defined in the capability and also restricted by capability constraints. In addition to these skill parameters which are connected to capability properties, there may also be skill parameters which are only relevant for execution and are thus not described on capability level.

The distinction between capabilities and skills decouples the description of a function from its implementation and enables developers to freely select a technology and programming language. In addition, multiple skill interfaces may be provided for one skill.
On the one hand, this allows developers to use existing and well-known technologies to offer a skill interface. On the other hand, integrators may select a skill interface matching their technology stack.

\section{Implementation of Model Elements}
\label{sec:application}
There is currently no standardization regarding the implementation of the \gls{css} reference model, and existing approaches typically favor different technologies for different model aspects. 
This section presents suitable implementation examples for the three model aspects capabilities, skills, and services.

\subsection{Modeling Capabilities using Ontologies}
\label{sec:CapabilityApplication}

Semantic Web technologies provide mechanisms for knowledge representation in information systems.
They are based on a stack of downward-compatible languages for information models and knowledge representation standardized by the W3C. Ontologies constitute reusable information models that capture the knowledge of a domain in a general form, independent of specific applications and are used as semantically rich schemas for knowledge graphs. The W3C technology stack for ontologies consists of the \gls{rdf}\footnote{\url{https://www.w3.org/RDF/}} and its Schema extension (RDFS)\footnote{\url{https://www.w3.org/TR/rdf-schema/}} that form the representational basis for the \gls{owl}\footnote{\url{https://www.w3.org/OWL/}}.
\gls{owl} allows to express domain knowledge in terms of logical statements that support automated reasoning for inferring implicit knowledge. 
Additional powerful technologies such as SPARQL\footnote{\url{https://www.w3.org/TR/rdf-sparql-query/}} and SHACL\footnote{\url{https://www.w3.org/TR/shacl/}} may be used to query for and validate the information in \gls{rdf}-based data models.

Semantic Web technologies provide an ideal candidate solution for the semantically rich representation of capabilities concerning their surrounding \gls{ppr} model elements and for the matching of semantic capability descriptions. A direct way of utilizing \gls{owl} is to model the notion of capability as an \gls{owl} class.
One can then use \gls{owl} restrictions on \gls{owl} properties for representing properties and their constraints from the \cssmodel. Applications can then introduce specific capabilities like \textsl{Drilling} by means of sub-classing together with their relevant properties restricted in complex \gls{owl} class expressions. An example of such an expression in \gls{owl} Manchester syntax is ``\texttt{Drilling and (depth some integer[<=15])}'' to represent a capability for \textsl{drilling with a depth of max. 15 mm}. The specialization hierarchies for capability classes can be taken from standards such as DIN 8580 or VDI 2860, as proposed in \cite{KHV+_AFormalCapabilityand_9820209112020}. This representational approach is similar to the one presented in \cite{WBS+_AnOntologybasedMetamodelfor_9820209112020} and extends it by equipping otherwise opaque capability classes with constraints on properties to account for their rich semantics.

Moreover, \gls{owl} Reasoning can be utilized for capability matching. The conjunction formed from two capability class expressions -- one offered by a resource and the other one requested for a process -- can be checked for satisfiability by any standard \gls{owl} reasoner to test whether the two capabilities are compatible, meaning that their constraint sets can be jointly fulfilled. This technique goes back to the intersection matching proposed in \cite{LiHo-www03}. As discussed in \cite{phd-grimm09}, issues with \gls{owl}'s open-world assumption can be overcome by strictly controlling the ontological vocabulary used.
This requires systematically including so-called closure axioms, which is presumably easy to achieve in a factory environment, e.g., disjointness between sibling capability classes from standard hierarchies.
Still, this approach needs further research on potential issues with scalability and expressivity in comparison to alternative constraint solving methods when applied in real setups with complex capability descriptions on large property sets.

\subsection{Executing Skills using OPC UA}
\label{sec:SkillApplication}

In recent years, the possibilities of implementing skills have been investigated in various publications and research projects, such as DEVEKOS, BaSys4.0/4.2, or AKOMI  \cite{MZG+_ChallengesinSkillbasedEngineering_2018,Dorofeev.2018,Zimmermann.2019,VeitHammerstingl.2019,Volkmann.2020b,Volkmann.2021}. 
The vendor-independent communication standard \gls{opc} has emerged as a promising approach for implementing skill interfaces. 
\gls{opc} has a high degree of diffusion in control technology due to its ability to provide a resource-neutral information model in its servers. 
The description of skills with all parameters can be mapped directly within this information model to enable unified control of resources. 
So far, standardized \gls{opc} information models (so-called "Companion Specifications"), focused primarily on data acquisition, e.g., for asset management or condition monitoring. 
These use cases require primarily \emph{read-only} access which does not enable complete interoperability of machines in the sense of the Industrie 4.0 vision. 
Furthermore, a classic real-time capable communication standard for controlling the resources is still required. 
This currently means there can either be separate networks for data acquisition and control or a shared network that strongly limits the available traffic. Therefore, it is essential to enable \emph{write} and, thus, control access to these machines and systems over \gls{opc}. Companion Specifications, such as the PackML state machine (OPC 30050) \cite{OPCFoundationPackML.2020d} or \gls{opc} programs (OPC 10000-10) \cite{OPCFoundationPart10.2017}, show first approaches for such access. However, there is a lack of a uniform and cross-domain concept for the realization of skills that \gls{opc} can provide. 

\begin{figure} [ht!]
  \begin{center}
  \includegraphics [width=\linewidth] {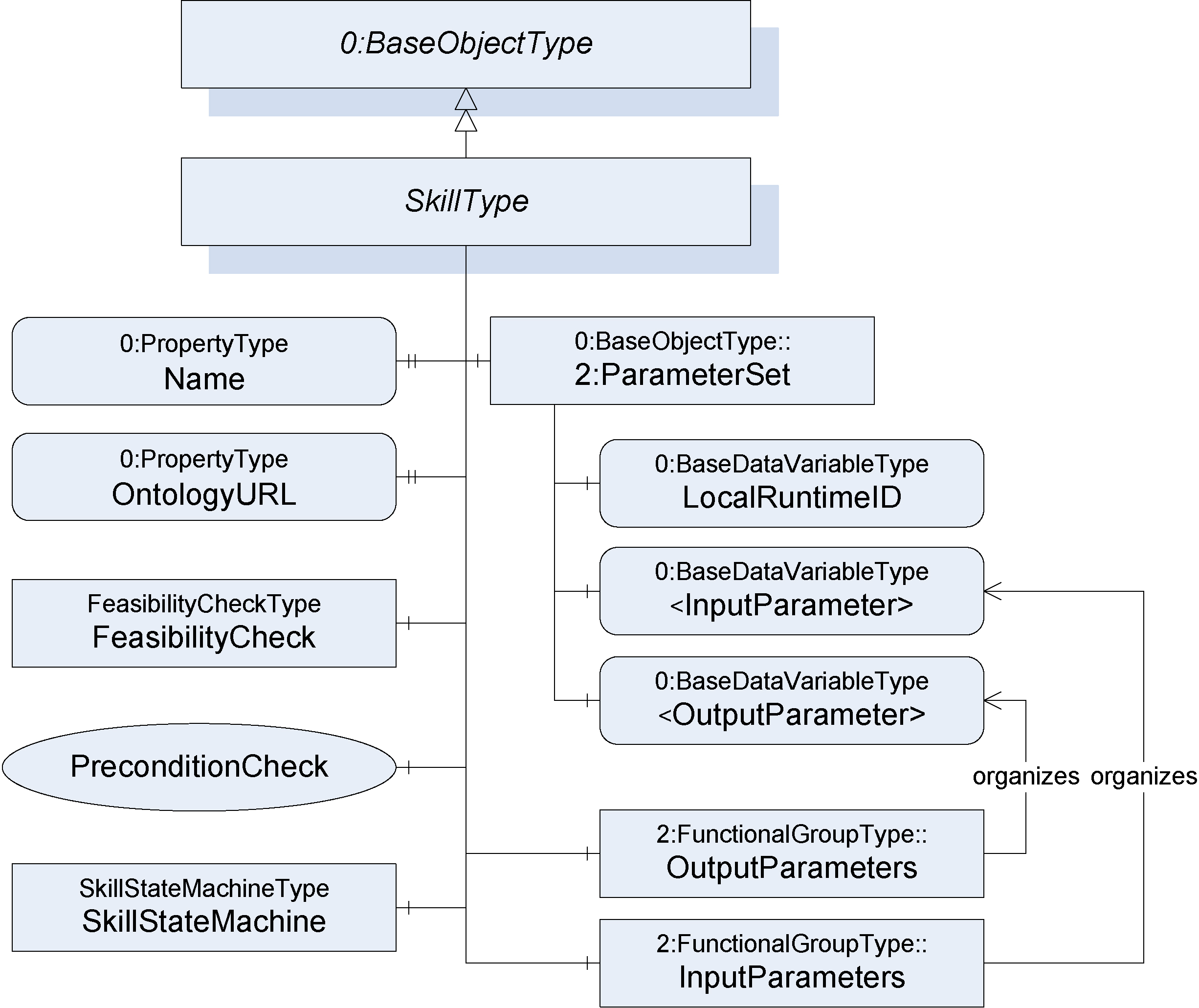}
  \caption{OPC UA Skill Metamodel}
  \label{fig:opc_ua_model}
  \vspace{-25 pt}
  \end{center}
\end{figure} 

As shown in Figure \ref{fig:reference_model}, a distinction between the skill interface and the actual skill is necessary. The skill interface is defined as an \gls{opc} information model. Figure \ref{fig:opc_ua_model} shows a proposal for this model.
A separate \gls{opc} ObjectType with the name ``SkillType'' is created, which provides the essential elements for the skill interface:
The \texttt{name} of the skill is optional and serves as plain text for the user to quickly identify the skill.
The \texttt{ontologyURL} must be specified and forms the reference to the capability in the ontology model (cf. paragraph \ref{sec:capabilities}). This reference can be used to identify the capability realized.
The \texttt{SkillStateMachine} represents a finite state machine similar to the aforementioned \gls{opc} for PackML or \gls{opc} Programs.
The \texttt{ParameterSet} contains necessary parameters to set or check during or after skill execution:
\begin{itemize}
\item The \texttt{LocalRuntimeID} serves a client as a unique identifier to identify the skill for execution. 
\item The placeholder \texttt{<InputParameter>} is used to define any input parameters necessary for the execution or configuration of a skill. These can be, for example, position or speed parameters.
\item The placeholder \texttt{<OutputParameter>} is used to define any output parameters that are returned by a skill. These can be, for example, current actual values (speed, velocity) during the skill execution, which are necessary for synchronization with other skills. Furthermore, it is also conceivable to create sensory skills, for example, for quality assurance, whose result is also represented by output parameters \cite{VeitHammerstingl.2019}. 
\end{itemize}

The optional \texttt{FeasibilityCheck} may be used to confirm the execution of complex skills in advance \cite{Volkmann.2021}. A feasibility check can also be implemented as a state machine in the \gls{opc} information model. To check the executability of a skill, the required input parameters are written into the parameter set of the feasibility check. 
The optional \texttt{PreconditionCheck} checks shortly before executing a skill whether the required resource fulfills all necessary conditions. This is especially needed if the execution depends on many other factors. In the assembly domain, this could be checking whether a required component is in stock or, in the field of machine tools, checking if needed tools are available. 
Furthermore, the concept of skills can be integrated into \gls{opc} PubSub over Time-Sensitive Networking (TSN) to realize real-time capable communication for skills.

\subsection{Describing Services using the Asset Administration Shell}
\label{sec:ServiceApplication}
There are few publications, such as \cite{Belyaev2018}, investigating the distinction of the concepts capability, skill, and service. However in existing publications, there are no implementation examples.
The exchange of information between companies must be based on standardized semantics. 
Therefore, one solution to describe services is the \gls{ass}. The \gls{ass} is an Industrie 4.0 specification of a digital twin to enhance interoperability between systems of different vendors. 
Recently published specifications define a standardized \gls{ass} meta-model and \gls{ass} interface \cite{AASSpecification}. 
The meta-model specifies a set of elements used to create I4.0-compliant information models, so-called Submodels (SM), needed to represent several aspects and functionalities of a modeled asset. Using standardized SMs to ensure cross-company interoperability includes the standardization of information models for describing various aspects of modeled assets. The service is either demanded by the ServiceRequester or offered by the ServiceProvider. Therefore, there must be two different SM templates. 
    
The request contains the specification of required product or process requirements as well as the description of the required provision and may be part of the product \gls{ass}. The description of the provision, relevant for potential manufacturers, may include necessary certifications and a non-disclosure agreement if required. 
The different categories can be organized in Submodel Collections (SMC) and can be described by properties. For instance, an SMC \emph{TenderCritera} may represent all the information needed to find a suitable manufacturer based on, e.g., the required quantities, price specifications, CO2 specifications, and delivery conditions. 
All SMEs and their properties are extended by a preset Qualifier, indicating that the information is a requirement. Each property can also be described by a semantic ID. In case of the service, which is used between companies, standardized data elements are required. Therefore repositories, such as IEC Common Data Dictionary\footnote{\url{https://cdd.iec.ch/cdd/iec61360/iec61360.nsf}} or ECLASS\footnote{\url{https://eclass.eu/eclass-standard/content-suche}}, are recommended to ensure common semantics.

The ServiceProvider offers services that can be matched to the required services. In this case, the SM service could be part of the factory or company \gls{ass}, based on a similar description as the requested service. The description is further extended by the capabilities which can provide the assured service to link the service and the corresponding capabilities.

\section{Alternative Approaches} 
\label{sec:alternativeApproaches}
Approaches to capabilities, skills and services are a rather new addition to a group of comparable research approaches. Even before the terms used in this paper appeared, some approaches pursued similar goals, i.e., the encapsulation of automation functions and the description of their interfaces.
In this section, a differentiation of the \gls{css} model with two comparable approaches is given.

\subsection{Web Services \& Service-oriented Architectures}
\label{sec:webservices}
Web Services are software systems that allow humans or machines to consume functionalities via a network. Interface descriptions in machine-readable formats are required for machine-to-machine interoperation. \cite{Booth:04:WSA}

A \gls{soa} is an architectural paradigm that encourages using multiple services to structure software functionality which may be distributed and maintained by different owners \cite{OASIS_SOARM_ReferenceModelforService}. Services are typically considered self-contained functions that may be composed of other services. They logically represent a recurring activity with a clearly defined input and output so that consumers of the service may interact with it in the sense of a ``black box'' --- i.e. without knowing a service's internal details \cite{The_TheSOASourceBook_2009}. 

The understanding of the term ``service'' as used in information technology differs significantly from the understanding expressed in this publication. While services in IT are encapsulated functionalities and may thus be compared to skills, services in the context of the \gls{css} model act as containers bundling capabilities with commercial aspects in order to be offered and requested on a marketplace (see Section~\ref{sec:services}).

In fact, transferring the \gls{soa} service concept from information technology to automation was one of the earliest approaches to obtaining encapsulated functions with clearly defined interfaces in automation \cite{JaSm_ServiceOrientedParadigmsinIndustrial_2005}. Thus, services can be seen as an early precursor of skills according to the \gls{css} reference model.

\subsection{Module Type Package}
A description of modular process units is provided with the \glspl{mtp}. The \gls{mtp} defines and describes data of the structure, information interfaces, process sequences, and functions (\gls{mtp}-Services) of modules from automation technology. The combination and aggregation of components enable modular process units, known as Process Equipment Assembly (PEA). A PEA is developed once and contains the physical design of the process step to be implemented as well as the information technology interface to higher-level systems. \cite{VDIVereinDeutscherIngenieuree.V..2019} 

A module description is contained in the \gls{mtp} to allow module integration into a modular process plant. Modules provide \gls{mtp}-services with a predefined behavior (procedure) and a standardized interface that are offered externally with their description as \gls{mtp}. 

\gls{mtp}-concepts  can be related to the beforementioned capabilities and skills, see \ref{sec:capabilities}). 
A description of an \gls{mtp}-service may be considered a capability. A skill --- being an executable implementation --- may be compared to the implemented \gls{mtp}-service with its procedure. 
A PEA is the described resource of the \gls{css} model, which provides skills in form of \gls{mtp}-services with its procedures and their implemented-independent description of a function as a capability. In contrast to the service of the \gls{css} model, the \gls{mtp}-service is at skill level and does not consider business, compliance or any commercial aspects.

Another relation to the concept of capability and module descriptions with its functions is given by the term ``Super-Service'' in \cite{Blumenstein.2019}. 
A ``Super-Service'' is described as a conceptual planning artifact, which contains the union of all process engineering services and its procedures. 
The objective is to break down module boundaries and describe them in a manner that is independent of modules or resources like capabilities.

\section{Conclusion \& Outlook} 
\label{sec:outlook}
Flexible production is a promising means to meet the challenges of fluctuating markets. 
A current starting point for this is to be able to automatically compare the functions required for the production of a product with the functions offered by the machines and plants during planning and operation. 
Such automatisms require that the components and stakeholders involved communicate these functions with the same understanding on all levels of detail. 
This paper proposes a CSS reference model that defines the concepts of capability, skill and service, and relates them to existing PPR concepts. 
The \gls{css} model reflects a notion of function on three levels of abstraction. It answers questions about the set of interconnected function-related concepts and their relationships in the context of flexible production. And it combines the mostly singular solution approaches available in the literature into a comprehensive reference model that enables interoperable solutions.
Initial examples for implementing the presented concepts were shown.

But the presented model is initially a conceptual model that was created as a result of projects and cross-organizational working groups under the umbrella of \emph{Plattform Industrie 4.0}.
Further research activities must be undertaken so that a comprehensive and consistent mapping of the model to selected technologies can be offered. A promising approach to use AAS has been started within the framework of the organization "Industrial Digital Twin Association" (IDTA).

\printbibliography

\end{document}